\documentclass[preprint]{article}

\usepackage[utf8]{inputenc} 
\usepackage[T1]{fontenc}    
\usepackage{hyperref}       
\usepackage{url}            
\usepackage{booktabs}       
\usepackage{amsfonts}       
\usepackage{nicefrac}       
\usepackage{microtype}      
\usepackage{amsmath}
\usepackage{graphicx}
\usepackage[algo2e]{algorithm2e}
\usepackage{algorithmic}
\usepackage{booktabs}
\usepackage{hyperref}
\usepackage{algorithm}
\usepackage{multirow}
\usepackage{xcolor}
\usepackage{enumitem}

\usepackage[nonatbib]{neurips_2021}
\usepackage[numbers]{natbib}

\newcommand{\given}{|}
\newcommand{\grad}{\nabla}
\newcommand{\bSigma}{\boldsymbol{\Sigma}}
\newcommand{\bmu}{\boldsymbol{\mu}}
\newcommand{\bL}{\mathbf{L}}

\title{Multivariate Probabilistic Regression with Natural Gradient Boosting}

\author{%
  Michael O'Malley \\
  STOR-i Centre for Doctoral Training\\
  Lancaster University, UK\\
  \texttt{m.omalley2@lancaster.ac.uk} \\
   \And
   Adam M. Sykulski \\
  Department Of Mathematics And Statistics\\
  Lancaster University, UK\\
  \And
  Rick Lumpkin\\
  NOAA/Atlantic Oceanographic and\\ Meteorological Laboratory, FL, USA \\
  \And
  Alejandro Schuler\\
  Department of Biomedical Informatics\\
  Stanford University, CA\\
}
\makeatletter \renewcommand\paragraph{\@startsection{paragraph}{4}{\z@}                                     {1mm}                                     {-1em}
{\normalfont\normalsize\bfseries}} \makeatother

\DeclareMathOperator{\Ex}{\mathbb{E}}

\begin{document}
\maketitle
\begin{abstract}
Many single-target regression problems require estimates of uncertainty along with the point predictions. Probabilistic regression algorithms are well-suited for these tasks. However, the options are much more limited when the prediction target is multivariate and a \textit{joint} measure of uncertainty is required. For example, in predicting a 2D velocity vector a joint uncertainty would quantify the probability of any vector in the plane, which would be more expressive than two separate uncertainties on the x- and y- components. To enable joint probabilistic regression, we propose a Natural Gradient Boosting (NGBoost) approach based on nonparametrically modeling the conditional parameters of the multivariate predictive distribution. Our method is robust, works out-of-the-box without extensive tuning, is modular with respect to the assumed target distribution, and performs competitively in comparison to existing approaches. We demonstrate these claims in simulation and with a case study predicting two-dimensional oceanographic velocity data. An implementation of our method is available at https://github.com/stanfordmlgroup/ngboost.
\end{abstract}

\section{Introduction}

The standard regression problem is to predict the value of some continuous target variable $\mathbf{Y}$ based on the values of an observed set of features $\mathbf{X}$. Under mean squared error loss, this is equivalent to estimating the conditional mean $\Ex[\mathbf Y|\mathbf X]$. Often, however, the user is also interested in a measure of predictive uncertainty about that prediction, or even the probability of observing any particular value of the target. In other words, one seeks to estimate $p(\mathbf{Y}|\mathbf{X})$. This is called \textit{probabilistic} regression.

Probabilistic regression has been approached in multiple ways; for example, deep distribution regression \citep{li2021deep}, quantile regression forests \citep{meinshausen2006quantile}, and generalised additive models for \textit{Location, shape, scale} \citep{rigby2019distributions}. All the listed methods focus on producing some form of outcome distribution. Both \citep{li2021deep} and \citep{meinshausen2006quantile} rely on binning the output space, an approach which will work poorly in higher output dimensions. Here we focus on a method closer to \cite{rigby2019distributions} where we prespecify a form for the desired distribution, and then fit a model to predict the parameters which quantify this distribution. Such a parametric approach allows one to specify a full multivariate distribution while keeping the number of outputs which the learning algorithm needs to predict relatively small. This is in contrast to deep distribution regression \cite{li2021deep} for example, where the number of outputs which a  learning algorithm needs to predict becomes infeasibly large as the dimension of the target data $\mathbf{Y}$ grows.

A common approach for flexible probabilistic regression is to use a standard multivariate regression model to parameterize a probability distribution function. This approach has been extensively used with neural networks \citep{williams1996using,sutzle2005numerical, rasp2018neural}. Recently \cite{NGBoost} proposed an algorithm called Natural Gradient Boosting (NGBoost) which uses a gradient boosting based learning algorithm to fit each of the parameters in a distribution using an ensemble of weak learners. A notable advantage of this approach is that it does not require extensive tuning to attain state-of-the-art performance. As such, NGBoost works out-of-the-box and is  accessible without machine learning expertise.
 
 \begin{figure}
     \centering
     \includegraphics[width=0.6\textwidth]{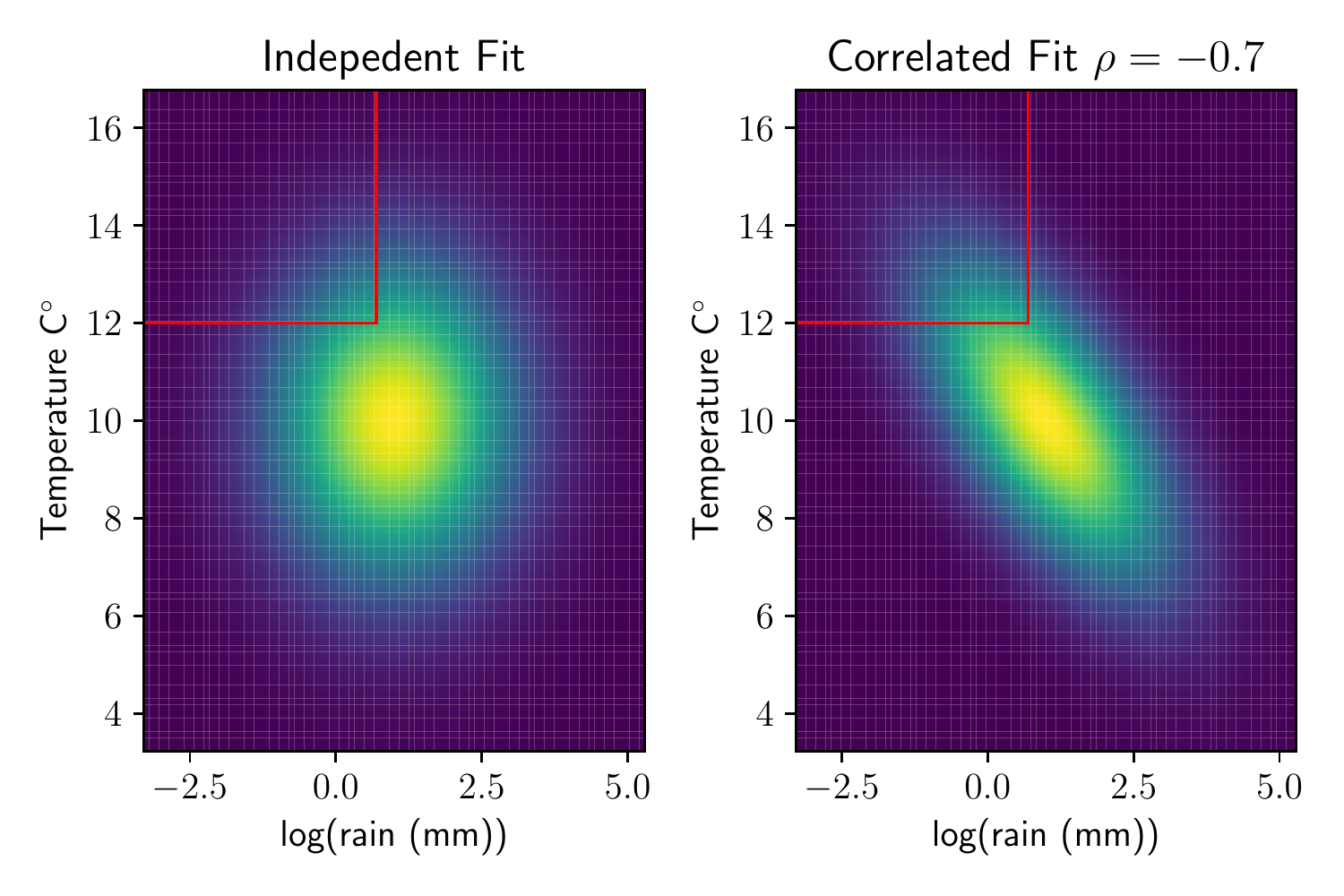}
     \caption{An example distribution of log rainfall and temperature for a day which we know will have rain. Both distributions are Gaussian and have the same mean vector and marginal variances. On the right plot the correlation between the two dimensions is set to -0.7, whereas on the left plot it is 0.}
     \label{fig:rain}
 \end{figure}
 
Our contribution is to construct and test a similar method that works for multivariate probabilistic regression. 
The standard approach for multi-output regression is to either fit each dimension entirely separately, or assume that the residual errors in both output dimensions are independent, allowing the practitioner to factor the objective function.
In reality, however, output dimensions are often highly correlated if they can be predicted using the same input data. When the task requires a measure of uncertainty, we must therefore model the \textit{joint} uncertainty in the predictions. 
 
As a simple illustrative example of where a joint uncertainty prediction is useful, consider the two joint distributions of rainfall and temperature as in Figure \ref{fig:rain}. Suppose these are the forecasts we have predicted and we are interested in the probability that it is going to rain at most 2mm, while the temperature does not go higher than $12^\circ$. Using the independent forecast we get a probability of 0.14; whereas using the fit which allows correlations between output dimensions, we get a probability of 0.25 based on the joint forecast. The benefits of joint probabilistic regression are clear: (1) the distribution allows one to make more accurate decisions based on both outcomes, and (2) the correlation measure is often a quantity of interest in itself (e.g., in forecasting stock prices for portfolio selection).
 
The motivating application of this paper is to predict two-dimensional velocity outputs. Such outputs are commonplace in environmental applications such as ocean currents \citep{maximenko2009mean, sinha2021estimating} and wind \citep{lei2009review}. In these approaches, a diagonal covariance matrix is often assumed; which will seldom be the case in practice. In Section \ref{sec:application}, we provide a real-world large scale example of using our approach to predict the multivariate distribution of ocean currents using multivariate remotely-sensed satellite data. 

In summary, in this paper we present an NGBoost algorithm that allows for multivariate probabilistic regression. Our key innovation is to place multivariate probabilistic regression in the NGBoost framework, which allows us to model all parameters of a \textit{joint} distribution with flexible regression models. At the same time, we inherit the ease-of-use and performance of the NGBoost framework. In particular, we demonstrate the use of a multivariate Gaussian NGBoost parametrization in simulation and in the use case described above. The results show that: (1) joint multivariate probabilistic regression is more effective than naively assuming independence between the outcomes, and (2) our multivariate NGBoost approach outperforms a state-of-the-art neural  network approach. Our code is freely available as part of the NGBoost python package.

\subsection{Notation}
Let the input space be denoted $\mathcal{X}=\mathbb{R}^d$ and the output space be denoted $\mathcal{Y}=\mathbb{R}^p$. The aim of this paper is to learn a conditional distribution from the training dataset $\mathcal{D} = \{\mathbf{X}_i, \mathbf{Y}_i \given 1\le i \le N~ \mathbf{X}_i \in \mathcal{X}, \mathbf{Y}_i \in \mathcal{Y}\}$. Let $p(\mathbf{Y}\given \boldsymbol{\theta})$ be a probability density of the observations $\mathbf{Y}\in \mathcal{Y}$ parameterized by $\boldsymbol{\theta}\in \mathbb{R}^M$, where $M$ is the dimension of the parameters. In this work we aim to learn a mapping such that the parameter vector varies for each data-point, i.e., a function $f:\mathcal{X} \rightarrow \mathbb{R}^M$, inducing the distribution function of $\mathbf{Y}$, $p\left(\mathbf{Y} \given \boldsymbol{\theta}=f(\mathbf{X})\right)$. For convenience we shall henceforth adopt a shorter notation: $p\left(\mathbf{Y} \given \boldsymbol{\theta}=f(\mathbf{X})\right)=p\left(\mathbf{Y} \given \mathbf{X}\right)$.

\section{Approach}
\label{sec:approach}
\subsection{Natural Gradient Boosting}
\label{ssec:ngboost}
Natural Gradient Boosting (NGBoost) \citep{NGBoost} is a method for probabilistic regression. The approach is based on the well-proven approach of boosting \citep{friedman2001greedy}. Rather than attempting to produce a single point estimate for $\Ex[\mathbf{Y} | \mathbf{X}]$, NGBoost aims to fit parameters for a pre-specified probability distribution function with $M$ unknown parameters, in turn producing a prediction for $p(\mathbf{Y} | \mathbf{X})$. Therefore the prediction this model produces includes a measure of aleatoric uncertainty for the data $\mathbf{Y}$. 

NGBoost learns this relationship by fitting a base learner $f^{(b)}:\mathcal{X} \rightarrow \mathbb{R}^{M}$ at each boosting iteration. The weighted sum of these base learners constitutes the function $f$. Each output of $f$ is used to parameterize the pre-specified distribution. We use $M$ independent model fits $f=\{f_m\}_{m=1}^M$ as the base learner \cite{NGBoost}. After $B\in \mathbb{N}$ boosting iterations, then the final sum of trees specifies a conditional distribution:
\begin{equation*}
    p\left(\mathbf{Y} {\large|} \boldsymbol{\theta}= \left\{\boldsymbol{\theta}^{(0)}-\eta\sum_{b=1}^B \rho^{(b)}f^{(b)}(\mathbf{X})\right\}\right),
\end{equation*}
where $\rho^{(b)}\in \mathbb{R},~b\in \{1,\dots,B\}$  are scalings chosen by a line search, $\boldsymbol{\theta}^{(0)}$ is the initialization of the parameters, and $\eta$ is the fixed learning rate.

The base learner at each iteration is fit to approximate the functional gradient of a scoring rule $\mathcal S$ (the probabilistic regression equivalent of a loss function). This takes the form of predicting the value of the gradient $\grad_{\boldsymbol \theta} \mathcal S(\mathbf Y, p_{\boldsymbol{\theta}}(\mathbf Y | \mathbf X))$ from the features $\mathbf X$ using a standard regression algorithm. In effect, each base learner represents a single gradient descent step. However, this approach by itself is not sufficient to attain good performance in practice. To solve problems with poor training dynamics, NGBoost fits using the natural gradient \citep{amari1998natural} in place of the ordinary gradient. This is particularly advantageous for fitting probability distributions as discussed in \cite{NGBoost}. 

To calculate the natural gradient, one must multiply the ordinary gradient by the inverse of a Riemannian metric $\mathcal{I}_{\mathcal{S}}(\mathbf{\theta})$ evaluated at the current point $\mathbf{\theta}$ in the parameter space. The nature of the Riemannian metric depends on the scoring rule $\mathcal S$ chosen to evaluate the predicted distribution against the observed data. In this paper we use the negative log-likelihood as the scoring rule which corresponds to maximum likelihood estimation. The relevant Riemannian metric in this case is the Fisher information matrix. For reference, we give pseudo-code in Algorithm~\ref{alg}.

\begin{algorithm}
\KwData{Dataset $\mathcal{D} = \{\mathbf{X}_i,\mathbf{Y}_i\}_{i=1}^N$.}
\KwIn{Boosting iterations $B$, Learning rate $\eta$, Probability distribution with parameter $\boldsymbol{\theta}$,  Scoring rule $\mathcal{S}$, Base learner $f$.}
 \KwOut{Scalings and base learners $\left\{\rho^{(b)},f^{(b)}\right\}_{b=1}^B.$ \vspace{0.02cm}}
$\boldsymbol{\theta}^{(0)} \gets \arg\min_{\boldsymbol{\theta}} \sum_{i=1}^{N}\mathcal{S}(\boldsymbol{\theta}, \mathbf{Y}_i)$ \algorithmiccomment{initialize to marginal}\;
\For{$b \gets 1,\hdots,B$}{
    \For{$i \gets 1,\hdots,N$}{
         $g_i^{(b)} \gets \mathcal{I}_{\mathcal{S}}\left(\boldsymbol{\theta}_i^{(b-1)}\right)^{-1}{\nabla_{\boldsymbol{\theta}}} \mathcal{S}\left(\boldsymbol{\theta}_i^{(b-1)}, \mathbf{Y}_i\right)$ \; 
    }
    
    $f^{(b)} \gets \mathsf{fit}\left(\left\{ \mathbf{X}_i, g_i^{(b)} \right\}_{i=1}^{{N}}\right)$ \;
    
    $\rho^{(b)} \gets \arg\min_\rho \sum_{i=1}^{{N}}\mathcal{S}\left( \boldsymbol{\theta}_i^{(b-1)} - \rho \cdot f^{(b)}(\mathbf{X}_i), \mathbf{Y}_i \right)$

     \For{$i \gets 1,\hdots,N$}{
        $\boldsymbol{\theta}_i^{(b)} \gets \boldsymbol{\theta}_i^{(b-1)} - \eta \left(  \rho^{(b)}\cdot f^{(b)}(\mathbf{X}_i)\right)$\;
    }
}
\caption{NGBoost for probabilistic prediction. Taken from \cite{NGBoost}} 
\label{alg}
\end{algorithm}

\subsection{A Multivariate Extension}
\label{ssec:mvn}

The original NGBoost paper \cite{NGBoost} briefly noted that NGBoost could be used to jointly model multivariate outcomes, but did not provide details. Here we show how NGBoost extends to multivariate outcomes and provide a detailed investigation of one useful parametrization, namely the multivariate Gaussian.

In univariate NGBoost ($p=1,~Y \in \mathbb{R}$), the predicted distribution is parametrized with $p(Y | \{\boldsymbol{\theta}_m = f_m(\mathbf{X})\})$ where $Y$ is a univariate outcome. For example, $Y|\mathbf X$ may be assumed to follow a univariate Gaussian distribution where $\mu$ and $\log\sigma$ are taken to be the parameter vector $\boldsymbol{\theta}$
(i.e., the output of NGBoost is $\boldsymbol\theta(\mathbf X) = (\mu(\mathbf X), \log \sigma(\mathbf X))$). Therefore to model a \textit{multivariate} outcome, all that is necessary is to specify a parametric distribution that has multivariate support, as we shall now show.

\paragraph{Multivariate Gaussian NGBoost} The multivariate Gaussian is a commonly used distribution and a natural choice for many applications. The rest of this paper is focused on the development and evaluation of a probabilistic regression algorithm using the multivariate Gaussian. Other multivariate distributions are also trivially accommodated by our framework as long as the natural gradient can be calculated with respect to some parametrization in $\mathbb R^M$. We leave the development and testing of alternatives to future work.

The multivariate Gaussian distribution is commonly written in the \emph{moment} parameterization as
\begin{equation*}
    \mathbf{Y}_i \sim  \mathbb{N}\left(\boldsymbol{\mu}, \boldsymbol{\Sigma} \right),
\end{equation*}
where $\boldsymbol{\Sigma}$ is the covariance matrix (a $p\times p$ positive definite matrix), and $\boldsymbol{\mu}$ is the mean vector (a $p\times 1$ column vector). Note that we only consider positive definite matrices for $\bSigma$ to ensure the inverse exists. We write the probability density function as
\begin{equation}
    p(\mathbf{Y}_i \given \boldsymbol{\mu}, \boldsymbol{\Sigma}) = (2\pi)^{-p/2}\boldsymbol|\boldsymbol{\Sigma}|^{-\frac{1}{2}} \exp\left[-\frac{1}{2}(\mathbf{Y}_i -\boldsymbol{\mu})^{T} \boldsymbol{\Sigma}^{-1} (\mathbf{Y}_i-\boldsymbol{\mu})\right].
\end{equation}

We fit the parameters of this distribution conditional on the corresponding training data $\mathbf{X}_i$ such that
$p(\mathbf{Y}_i\given \mathbf{X}_i) = p(\mathbf{Y}_i\given \boldsymbol{\mu}(\mathbf{X}_i), \boldsymbol{\Sigma}(\mathbf{X}_i))$.
To perform unconstrained gradient based optimization for any distribution, we must have a parameterization for the multivariate Gaussian distribution where all parameters lie on the real line. The mean vector $\boldsymbol{\mu}$ already satisfies this. However, the covariance matrix does not; it lies in the space of positive definite matrices. We shall model the inverse covariance matrix which is also constrained to be positive definite. We leverage the fact that every positive definite matrix can be factorized using the Cholesky decomposition with positive entries on the diagonals \citep{BanerjeeSudipto2014LAaM}.

We opt to use an upper triangular representation of the square root of the inverse covariance matrix $\boldsymbol{\Sigma}^{-1} = \mathbf{L}^\top \mathbf{L}$, as used by \cite{williams1996using},
where the diagonal is transformed using an exponential to force the diagonal to be positive. As an example, in the two-dimensional case we have
\begin{equation*}\mathbf{L} = \left[\begin{matrix}
\exp(a_{11})& a_{12} \\
0 & \exp(a_{22})
\end{matrix}\right],\end{equation*} which yields the inverse covariance matrix
\begin{equation*}
    \boldsymbol{\Sigma}^{-1} = \left[\begin{matrix}
\exp(a_{11})^2& \exp({a_{11}})a_{12} \\
\exp(a_{11})a_{12} & a_{12}^2+\exp(a_{22})^2
\end{matrix}\right].
\end{equation*}
This parameterization for $\mathbf{\Sigma}^{-1}$ ensures that the resulting covariance matrix $\bSigma=(\bL^\top\bL)^{-1}$ is positive definite for all $a_{ij} \in \mathbb{R}$. Hence, we can fit the multivariate Gaussian in an unconstrained fashion using the parameter vector $\boldsymbol{\theta} = (\mu_1, \mu_2, a_{11}, a_{22}, a_{12})$ as the output in the two-dimensional case. Note that the number of parameters grows quadratically with the dimension of the data. Specifically, the relation between $M$, the dimension of $\boldsymbol{\theta}\in \mathbb{R}^M$, and $p$, the dimension of $\mathbf{Y}_i$, is
\begin{equation}
    M = \dfrac{p^2+3p}{2}.
\end{equation}

For NGBoost using the log-likelihood scoring rule, we require both the gradient and the Fisher Information. The gradient calculations are given in \cite{williams1996using}, and the derivations for the Fisher information are given in the supplementary information. We have used these derivations to add the multivariate Gaussian distribution to the open-source python package NGBoost \citep{NGBoost} as part of this paper. Note that the natural gradient is particularly advantageous for multivariate problems such as these. This is because, for example, there are multiple equivalent parameterizations for the multivariate Gaussian distribution \citep{sutzle2005numerical, salimbeni2018natural, malago2015information}, but the choice of parameterization has been shown to be less important when using natural gradients than with classical gradients (see, e.g. \cite{salimbeni2018natural}).

\section{Simulation}\label{sec:Simulation}

We now demonstrate the effectiveness of our multivariate Gaussian NGBoost algorithm in simulation. Specifically, we show that (a) it outperforms a naive baseline where the target components are modeled independently, (b) it outperforms a state-of-the art neural network approach, and (c) the natural gradient is a key component in the effective training of distributional boosting models in the multivariate setting.

\begin{figure}[ht]
    \centering
    \includegraphics[width=0.9\linewidth]{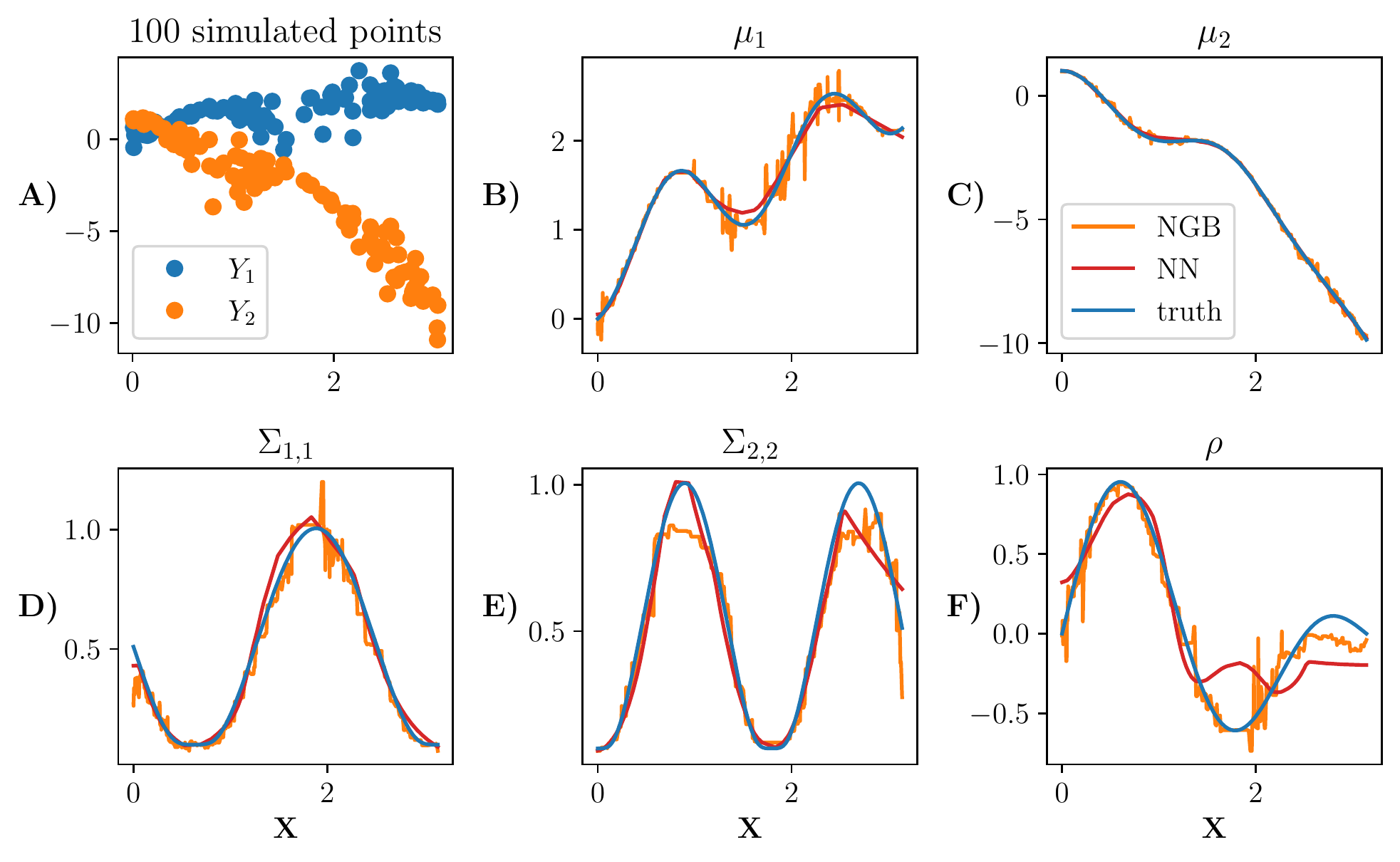}
    \caption{Simulated data from equation \eqref{eq:sim_eq}. A sample of 100 points is shown in A). We plot each parameter in plots B) to F). The true parameters of the distribution are shown in blue, the NGBoost fit is shown in orange, and the neural network fit with one hidden layer (100 neurons in the hidden layer) is shown in red. Both model fits are trained on $N=5000$ training points.}
    \label{fig:sim_data}
\end{figure}

We simulate the data similarly to \cite{williams1996using}. The nature of the simulation tests each algorithm's ability to uncover nonlinearities in each of the distributional parameter's relationship with the input. Specifically, we use a one-dimensional input and two-dimensional output, allowing us to illustrate the fundamental benefits of our approach in even the simplest multivariate extension.
The data are simulated as follows:
\begin{align*}
    \mathbf X_i &\overset{\text{IID}}{\sim} \text{Uniform}(0, \pi) \quad i \in \{1, \dots, N\}\\
    \mathbf Y_i | \mathbf X_i &\sim \mathbb{N} \left( \left[\begin{array}{c}
        \mu_1(\mathbf X_i)  \\
          \mu_2(\mathbf X_i)
    \end{array} \right], \left [  \begin{array}{cc}
        \sigma_{1}^2(\mathbf X_i) &\sigma_1(\mathbf X_i)\sigma_2(\mathbf X_i) \rho(\mathbf X_i)  \\
     \sigma_1(\mathbf X_i)\sigma_2(\mathbf X_i)\rho(\mathbf X_i) & \sigma_2^2(\mathbf X_i)
    \end{array} \right ] 
    \right),
\end{align*}
with the following functions:
\begin{align}
    \mu_1(x) &= \sin(2.5x)\sin(1.5x)+x, &  \mu_2(x) &= \cos(3.5x)\cos(0.5x)-x^2,\nonumber \\
    \sigma_1^2(x) &= 0.01 + 0.25[1-\sin(2.5x)]^2,&  \sigma_2^2(x) &= 0.01 + 0.25[1-\cos(3.5x)]^2, \nonumber\\
    \rho(x) &= \sin(2.5x)\cos(0.5x).\label{eq:sim_eq}
\end{align}
Simulated data alongside the true parameters are shown in Figure \ref{fig:sim_data}. 

We compare multiple methods, all of which predict a multivariate Gaussian. For all boosting models we use scikit-learn decision trees \citep{pedregosa2011scikit} as the base learner. Specifically, we consider five different comparison methods:\footnote{Software packages: NGBoost, tensorflow and scikit-learn are all available under an Apache 2.0 License.}
\begin{itemize}[leftmargin=*]
    \item \textbf{NGB:} The method proposed in this paper: Natural gradient boosting to fit a multivariate Gaussian distribution.
    \item \textbf{Indep NGB:} Independent natural gradient boosting where a univariate Gaussian model  is fitted for the two dimensions separately, i.e., $\mathbf{Y_i}\sim (\mathbb{N}(\mu_1, \sigma_2), \mathbb{N}(\mu_2, \sigma_2))$. Early stopping allows the number of trees used to predict each dimension to differ.
    \item \textbf{skGB:} scikit-learn's \citep{pedregosa2011scikit} implementation of gradient boosting (skGB) is used as a point prediction approach. To turn the skGB predictions into a multivariate Gaussian, we estimate a constant diagonal covariance matrix based on the residuals of the training data. For metric computation we assume that each $\mathbf{Y}_i$ follows a multivariate Gaussian with mean from a model fit for each dimension, and constant covariance matrix. Similar to Indep NGB we allow a different number of trees for each dimension.
    \item \textbf{GB:} A multi-parameter gradient boosting algorithm, where the only change from NGB is that the gradients are not multiplied by $\mathcal{I}_{\mathcal{S}}(\boldsymbol{\theta}_i^{(m-1)})^{-1}$.
    \item \textbf{NN:} The neural network approach from \citep{williams1996using}. We fit the multivariate Gaussian using tensorflow \citep{tensorflow}. We run a grid search on the network structure, as explained in the supplementary information.
\end{itemize}

To prevent overfitting, we employ early stopping for all methods on the held-out validation set, where we use a patience of 50 for all methods. For all boosting approaches we use the estimators up to the best iteration that was found, and for neural networks we restore the weights back to the best epoch. The base learner for all boosting approaches are run using the default parameters specified in each package. For each $N$ considered in the experiment, the neural network structure with the best log-likelihood metric averaged over all replications is chosen from the grid search. Note that we do not retrain the models after selecting the early stopping epochs/number of learners. We found that doing so gave a large advantage to the boosting based approaches. A learning rate of $0.01$ is used for all methods, for NN we use the Adam optimizer \cite{kingma2014adam}.

We ran the experiment with $N \in \{500, 1000, 3000, 5000, 8000, 10000\}$ with $50$ replications for each value of $N$. For each simulation, $N$ values were simulated as the training dataset, $300$ values were simulated as the validation set, and $1000$ values were used as the test set. 

The per-model results on the test data points are shown in Table \ref{tbl:metrics}. The average Kullback-Leibler (KL) divergence from the predicted distribution to true distribution is used as a metric. The results show that the NGB method is best for all $N$ despite not being tuned in any way besides early stopping. The KL divergence converges quickly to zero as $N$ grows, showing that NGB gets better at learning the distribution as $N$ increases. NN performs worse than NGB with higher KL divergence for all values of $N$. The KL divergence for Indep NGB seems to converge to a non-zero quantity as $N$ increases, likely because the target distribution in equation \eqref{eq:sim_eq} cannot be captured by a multivariate normal with a diagonal covariance matrix. skGB has large KL divergences which get worse as $N$ increases, likely because of the homogeneous variance fit. We note that GB performs significantly worse than NGB showing that the natural gradient is necessary to fit the multivariate Gaussian effectively with boosting. Further metrics can be seen in the supplementary information, in particular they show GB fits the mean $\boldsymbol{\mu}$ very poorly, which is then compensated for by a large covariance estimate, explaining the large KL divergence.

The modification we made from the simulation in \cite{williams1996using} is that we added $x$ and $-x^2$ terms to $\mu_1(x)$ and $\mu_2(x)$ respectively in equation \eqref{eq:sim_eq}. This modification is to highlight where our method excels. We also ran the original simulation without the modification, where the results are given in the supplementary information. As a summary we note the major differences: (1) the gaps between NGB, GB and NN are smaller, and (2) the NN method does best for $N \in \{500, 1000, 3000\}$, NGB does best for $N\in \{5000, 8000, 10000\}$.
\begin{table}[ht]
 \caption{Average KL divergence in the test set (to 3 decimal places) from the predicted distribution to the true distribution as the number of training data points $N$ varies. Standard error estimated from 50 replications reported after $\pm$ to 3 decimal places. Lower values are better, the result with the lowest mean is bolded in each row. An extended table showing additional metrics is shown in the supplementary information.}
    \label{tbl:metrics}
    \centering
\begin{tabular}{llllll}
\toprule
N &            NGB &      Indep NGB &            skGB &               GB &             NN \\\toprule
500   &  \textbf{0.564±0.016} &  1.633±0.043 &  17.194±0.301 &  126.227±2.579 &  1.285±0.547 \\
1000  &  \textbf{0.257±0.004} &    1.150±0.020 &   17.963±0.270 &  114.113±1.622 &   0.320±0.018 \\
3000  &  \textbf{0.106±0.002} &  0.884±0.015 &  19.609±0.248 &   97.682±1.397 &  0.149±0.004 \\
5000  &  \textbf{0.081±0.008} &  0.878±0.015 &  20.308±0.169 &   90.101±1.291 &  0.128±0.005 \\
8000  &  \textbf{0.053±0.004} &  0.866±0.013 &   20.614±0.170 &   79.006±1.168 &  0.103±0.004 \\
10000 &  \textbf{0.043±0.001} &   0.831±0.010 &   20.554±0.150 &   74.799±1.191 &   0.130±0.004 \\
\bottomrule
\end{tabular}

\end{table}

\section{Real Data Application}\label{sec:application}
The motivating application of this paper is to predict two-dimensional oceanographic velocities from satellite data on a large spatial scale. Typically, this is done through physics-inspired parametric models fitted with least-squares or similar metrics, treating the directional errors as independent \citep{mulet2021new}.
\subsection{Data}
\label{ssec:Data}
Here we introduce the datasets used which shall define $\mathbf{Y}_i\in \mathbb{R}^2$, $\mathbf{X}_i\in \mathbb{R}^9$. The data for all sources is available from 1992 to 2019 inclusive. All data sources are publicly available. 

For the model output $\mathbf{Y}_i$, we seek to predict two-dimensional oceanic near-surface velocities as functions of global remotely-sensed climate data. The dataset used to train, validate and test our model comes from the Global Drifter Program, which contains transmitted locations of freely drifting buoys as they drift in the ocean and measure near-surface ocean flow. The quality controlled 6-hourly product is used \citep{GDP} to construct the velocity observations. We drop observations which have a high location noise estimate, and we low-pass filter the velocities at 1.5 times the inertial period (a timescale determined by the Coriolis effect), following previous similar works \citep{laurindo2017improved}. We only use data from buoys which still have a drogue (sea anchor) attached and thus more accurately follow ocean near-surface flow. We use the inferred two-dimensional velocities of these drifting buoys as our outputs $\mathbf{Y}_i$. 

The longitude-latitude locations of these observations and the time of year (percentage of 365) are used as three of the nine features in $\mathbf{X}_i$ to account for spatial and seasonal effects. For the remaining six features in $\mathbf{X}_i$, we use two-dimensional longitude-latitude measurements of \emph{geostrophic velocity} ($ms^{-1}$), \emph{surface wind stress} ($Pa$) and \emph{wind speed} ($m s^{-1}$). These variables are used as they jointly
capture geophysical effects known as geostrophic currents, Ekman currents, and wind forcing, which are known to drive oceanic near-surface velocities. We obtain geostrophic velocity from \cite{ssh} and both wind related quantities from \cite{wind}, where the data are interpolated to the longitude-latitude locations of interest. We further preprocess the data as explained in the supplementary information. To allow us to run multiple model fits, we subset the data to only include data points which are spatially located in the North Atlantic Ocean as defined by IHO marine regions \cite{IHO} and between $83^\circ \text W$ and $40^\circ \text W$ longitude. We use a temporal gridding of the buoy data of one day. This results in $415451$ observations for each input and output variable in the combined data set used for training, validating and testing our probabilistic regression model.

\subsection{Metrics} \label{ssec:metrics}
We cannot use KL divergence as in the simulation of Section \ref{sec:Simulation} because the true distribution of these data is unknown. Therefore we use a series of performance metrics that diagnose model fit.

\textbf{Negative Log-Likelihood (NLL):} All methods are effectively minimising the negative log-likelihood; therefore we use negative log-likelihood as one of our metrics: \[1/N \sum_i \log p(\mathbf{Y}_i\given \mathbf{X}_i).\]

\textbf{RMSE:} To compare the point prediction performance of the models we also report an average root mean squared error (RMSE): \[\left[ 1/(Np)\sum_{i=1}^N \sum_{j=1}^p (\mathbf{Y}_{i,j} - \hat{\mathbf{Y}}_{i,j})^{2}\right]^{1/2}, \quad \text{where~} \hat{\mathbf{Y}}_i = \Ex[\mathbf{Y}\given \mathbf{X}_i].\]

\textbf{Region Coverage and Area:}
As this paper focuses on a measure of probabilistic prediction, we also report metrics related to the prediction region. The prediction region is the multivariate generalization of the one-dimensional prediction interval. The two related summaries of interest are: (1) the percentage of $\mathbf{Y}_i$ covered by the $\alpha\%$ prediction region, and (2) the area of the prediction region.

A $\alpha\%$ prediction region can be defined for the multivariate Gaussian distribution as the set of values $\mathbf{Y}$ which satisfy the following inequality:
\begin{equation}
    (\mathbf{Y}-\boldsymbol{\mu})\boldsymbol{\Sigma}^{-1}(\mathbf{Y}-\boldsymbol{\mu})^\top \le \chi^2_{p,\alpha},\label{eq:CI}
\end{equation}
where $\chi^{2}_{p, \alpha}$ is the  quantile function of the $\chi^2$ distribution with $p$ degrees of freedom evaluated at $\alpha\%$.
This prediction region forms a hyper-ellipse which has an area given by
\begin{equation}
    \dfrac{(2\pi)^{p/2}}{p\Gamma(\frac{p}{2})} (\chi^2_{p,\alpha})^{p/2}|\boldsymbol{\Sigma}|^{1/2}.\label{eq:AREA}
\end{equation}
We report the percentage of data points which satisfy equation \eqref{eq:CI}, and we report the average area of the prediction region over all points in the test set.

\subsection{Numerical Results} \label{ssec:num_res}
	We compare the same five models considered in Section~\ref{sec:Simulation}. To compare the models we randomly split the dataset, keeping each individual buoy record within the same set. We put 10\% of records into the test set, 9\% of the records into the validation set, and 81\% into the training set. The model is fitted to the training set with access to the validation set for early stopping, and then the metrics from Section~\ref{ssec:metrics} are evaluated on the test set. This procedure is repeated 10 times.
 
    For this example we also use a grid search for all of the boosting approaches, in addition to the neural network approach. We found that with the default parameters the boosting methods generally resulted in under-fitting. For each boosting method, a grid search is carried out over number of leaves, and minimum data in leaves, as outlined in the supplementary material. The best hyper-parameters from the grid search for each method are selected by evaluating the test-set negative log-likelihood.
    
\begin{figure}[ht]
\includegraphics[width=0.9\textwidth]{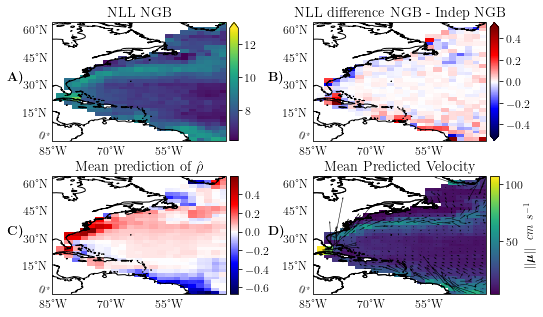}
\caption{Summaries of test set results within $2^\circ\times 2^{\circ}$ latitude-longitude bins for the North Atlantic Ocean application. A) shows the spatial distribution of the negative log-likelihood for NGB, B) shows the difference between the negative log-likelihood spatially between NGB and Indep NGB; negative values (red) implying NGB is better than Indep NGB (with vice versa in blue).
C) shows the average prediction of $\rho$ where $\rho =\boldsymbol{\Sigma}_{0,1}/\sqrt{ \boldsymbol{\Sigma}_{0,0}\boldsymbol{\Sigma}_{1,1}}$ is extracted from the predicted covariance matrix in the held out set from NGB. D) shows the mean currents estimated by NGB. All major ocean features are captured by the model \cite{lumpkin2013global}.}
\label{fig:spatial_metrics}
\end{figure}

   \begin{table}[ht]
    \centering
    \caption{
Average test set metrics defined in Section \ref{ssec:metrics}. Standard error estimated from 10 replications reported after $\pm$. PR stands for prediction region. Coverage has been shortened to cov. All numbers rounded to 2 decimal places, aside from the 90\% PR Area row which is rounded to the nearest integer. Lower values are better for NLL and RMSE, best value is bolded in both rows. Coverage and area must be considered in combination, for an ideal prediction the area would be low and the coverage would be 0.9.}
    \label{tab:results}
\begin{tabular}{llllll}
\toprule
 Metric &            NGB &      Indep NGB &           skGB &              GB &              NN \\
\midrule
NLL      &    \textbf{7.73±0.02} &    7.74±0.02 &   8.17±0.02 &    8.79 ± 0.01 &    7.81±0.02 \\
RMSE {\tiny$cm~s^{-1}$}     &   14.53±0.14 &   14.45±0.12 &  \textbf{14.31±0.13} &   24.14±0.36 &   16.63 ± 0.19 \\
90\% PR cov &     0.87±0.00 &     0.87±0.00 &    0.89±0.00 &     0.94±0.00 &     0.89±0.00 \\
90\% PR area  {\tiny$cm^{2}~s^{-2}$}   &  2482 ± 51 &  2568±41 &  2714±8 &  8070±22 &  3670±75 \\
\bottomrule
\end{tabular}

\end{table}
    The aggregated results of the model fits are shown in Table \ref{tab:results}. NGB and Indep NGB perform very similarly in terms of NLL, RMSE and 90\% PR coverage in this example. However, NGB provides a smaller 90\% PR area, which is expected as correlation will reduce $|\bSigma|$ in equation \eqref{eq:AREA}. To highlight the differences further, we show the spatial differences in negative log-likelihood between these two methods in Figure \ref{fig:spatial_metrics}B). In Figures \ref{fig:spatial_metrics}C) and D) we show the averaged held out spatial predictions for $\rho$ and $\mu$ from the NGB model. The results shown in $B)$ and $C)$ suggest that model stacking may be suitable to this application. For example, in geographic areas with high anticipated correlation use NGB; otherwise use Indep NGB.
    
    In Table \ref{tab:results} we see that the NN and GB approaches do poorly overall. Both methods have large root mean-squared error, which is compensated for by a larger prediction region on average, as can be concluded from the large average PR area. This behavior agrees with the univariate example given in Figure 4 of the original NGBoost paper \citep{NGBoost}, and the behavior of GB in the simulation of Section \ref{sec:Simulation}.

\section{Conclusions}
 \paragraph{Limitations}
 This paper has demonstrated the accuracy of our NGBoost method when focusing on bivariate outcomes with a multivariate Gaussian distribution. The derivations of the natural gradient and implementation in NGBoost are supplied for any dimensionality, but empirical proof of performance in higher dimensions is left to future work. Due to the quadratic relationship between $p$ and the number of parameters used to parameterize the multivariate Gaussian, the complexity of the learning algorithm greatly increases with larger values of $p$. Investigating a reduced rank form of the covariance matrix may be of interest for higher $p$ to reduce the number of parameters that need to be learned. We also leave the development and testing of alternative multivariate distributions to future work. The modular nature of our implementation makes it easy for users to experiment and add their own multivariate distributions as long as they can supply the relevant gradient and Riemannian metric functions.
 
The difference between Indep NGB and NGB is small in our real-world application, although NGB performed relatively much better in our simulation study. Generally, one should anticipate NGB to excel in cases with high correlation between the outcomes, whereas a method assuming independence should suffice when that assumption is warranted. Deep learning approaches are warranted in cases where a neural network naturally handles the input space such as images, speech or text.
 
\paragraph{Final Remarks} 
 In this paper we have proposed a technique for performing multivariate probabilistic regression with natural gradient boosting. We have implemented software in the NGBoost Python package which makes joint probabilistic regression easy to do with just a few lines of code and little to no tuning. Our simulation and case study show that multivariate NGBoost meets or exceeds the performance of existing methods for multivariate probabilistic regression.

\bibliography{mvn_pred}

\appendix
\renewcommand\thefigure{S\arabic{figure}}    
\setcounter{figure}{0}  
\renewcommand\thetable{S\arabic{table}}    
\setcounter{table}{0}  
\renewcommand\theequation{S\arabic{equation}}    
\setcounter{equation}{0}  

\section{Probabilistic Neural Networks}
In the paper, an existing neural network approach (referred to as NN) is used to fit conditional multivariate Gaussian distributions. The method fits a neural network taking inputs $\mathbf{X} \in \mathcal{X}$ where the output layer has $M$ units, which are used as $\boldsymbol{\theta}$ in the probability density function parameterization in Section 2.2 \cite{williams1996using, sutzle2005numerical}. The loss function used when optimizing the neural network parameters is the negative log-likelihood.

Instead of fitting $M$ independent models as in Section 2.1, when using probabilistic neural networks we fit one model which has $M$ outputs, one for each parameter of the probability density function. Throughout the paper, a fully connected neural network structure is used. We describe the structure search in Section B of this document.

Note that, unlike NGBoost, the natural gradient is not used in the optimization of the probabilistic neural networks to account for the geometry of the distribution space. 

\section{Machine Learning Model Details}
We cease the model fit when the validation score has not improved for 50 iterations. For the neural network approach, an iteration is one \emph{epoch} which is a full pass of the training dataset. For boosting approaches, an iteration is the growth of one base learner $f^{(m)}$ as explained in Section 2.1. If early stopping does not occur, we specify a maximum of 1000 iterations, however in our results this was only reached in the application section when using the GB method. The validation score used is the same as the scoring rule or loss for the method, i.e., root mean squared error for skGB, negative log-likelihood for all other methods.

\paragraph{Boosting Grid Search}
In our reported results, the boosting methods (NGB, GB, skGB, and Indep NGB) all used the same base learner, an sklearn decision tree, resulting in a direct mapping between hyper-parameters. Hence, we carried out the same grid search for these four models. We conducted a grid search over the following sets:
\begin{itemize}
	\item Max depth in $\{8, 15, 31, 64\}$.
	\item Minimum data in leaf in $\{1,15, 32\}$.
\end{itemize}
which resulted in 12 hyper-parameter sets for each method, and all other parameters were left at the default other than the learning rate. We did not tune the learning rate, we used a learning rate of 0.01 for the simulation and 0.1 for the application. The larger learning rate was chosen for the application to allow us to conduct replicated fits in a reasonable amount of time.

\paragraph{Neural Network grid search} 
We carried out the following grid search for the neural network architecture in both the simulation example and application in Sections 3 and 4, respectively. We considered the following structures for the hidden layers in the grid search:
\begin{itemize}
	\item A 20 unit layer
	\item A 50 unit layer
	\item A 100 unit layer (best for simulation when $N\in\{1000,3000,5000,10000\}$)
	\item A 20 unit layer followed by another 20 unit layer 
	\item A 50 unit layer followed by another 20 unit layer. (best for simulation when $N\in \{500,8000\}$)
	\item 100 units followed by another 20 unit layer (best for the application). 
\end{itemize}
After each layer, we applied a RELU activation on each node aside from the output nodes. A fixed batch batch size of 256 was used. For the simulation, a learning rate of 0.01 worked well for the Adam optimizer. In the application, we added an extra parameter into the grid search, a learning rate of 0.001 and 0.01. Adam's other parameters were left as the defaults in tensorflow. For the application, the best learning rate and structure pair was 0.001 and a 100 unit layer followed by a 20 unit layer respectively. 

\paragraph{Timings \& Computation}
All model fits were run on an internal computing cluster using only CPUs. We note that computational time was not a key consideration in this work. If it was of key importance, using a more efficient base learner for all boosting based approaches would be advisable for large $N$, such as LightGBM \cite{ke2017lightgbm} or XGBoost \citep{chen2016xgboost}. 
\begin{figure}
	\centering
	\includegraphics[width=\textwidth]{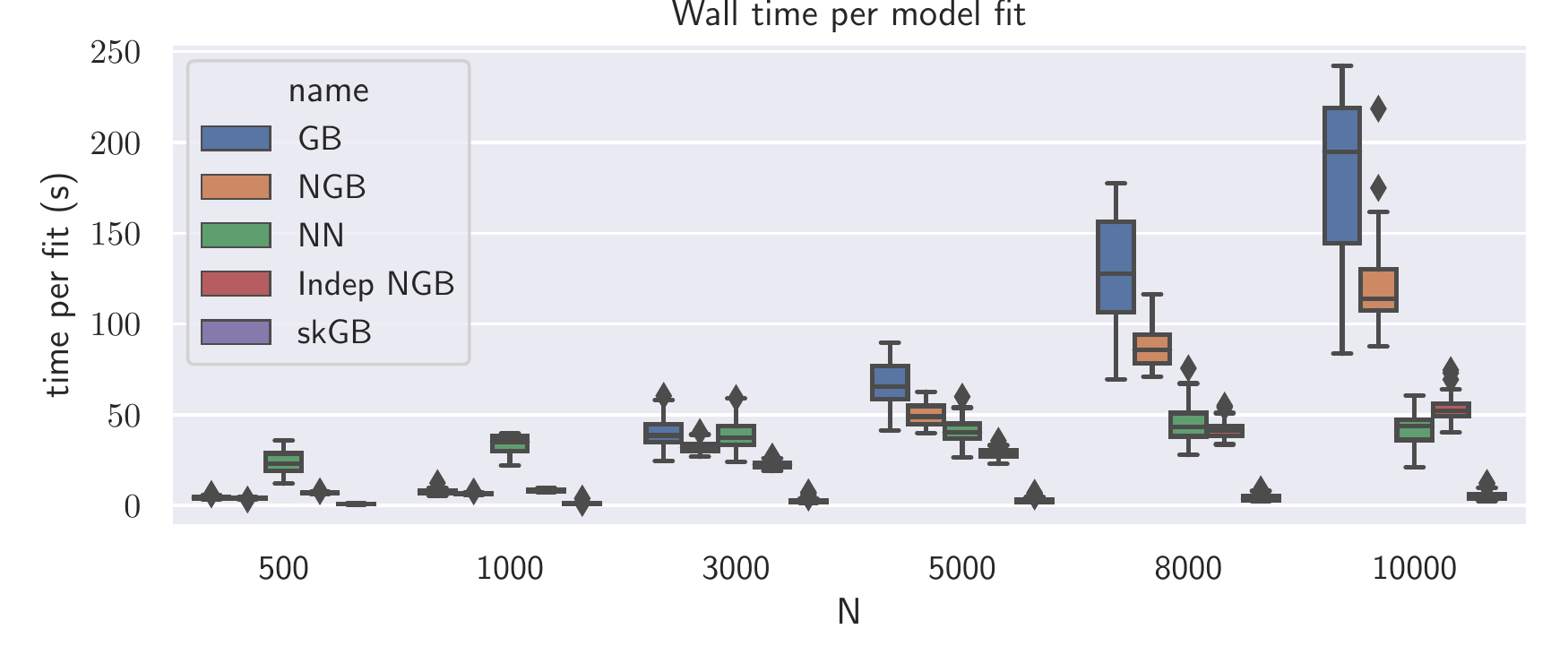}
	\caption{Wall Times per model fit over the 50 replications of Section 3. Box plot drawn in the standard way (boxes being a 50\% inter-quartile range, with lines drawn at median and at the minimum or maximum excluding `outliers`). NN model fits were given access to 10 threads, all other methods ran on a single thread. The plotting times are for a single model fit. It does not show the times relating to the grid search carried out for the NN method.}
	\label{fig:sim_timings}
\end{figure}

\textbf{Simulation}
We show the wall times for the simulation which we ran for Section 3 in Figure \ref{fig:sim_timings}. Generally, NGB and GB are the slowest and skGB is fastest. 

\textbf{Application}
We did not explicitly time each model fit in the application, hence we only give rough estimates based on the total running time for the best selected models from the grid search. The neural networks were given access to 15 CPU threads. All other methods were limited to 1 CPU thread, the difference between wall time and CPU time was negligible for all boosting approaches, hence we only report the wall time for those here. The estimated fitting times are shown below (reported as total time for the 10 random replications divided by 10):
\begin{itemize}
	\item NGB: 1.45 hours
	\item GB: 6 hours
	\item skGB: 1.3 hours
	\item NN: 0.7 hours wall time, 2.1 hours CPU time.
\end{itemize}
The GB method was the slowest method. This is because early stopping did not occur, meaning the algorithm would run for the full 1000 boosting iterations. In contrast, for NGB, early stopping usually occurred around 250 boosting iterations, which reduced the run time. 

We reiterate that these times are rough estimates. Both the simulation and the application were run on a cluster with various nodes with shared resources which the model fitting would be assigned to. For the timings shown in the list above, the NN approach was fit on a node with a 2.6GHz AMD Opteron Processor 6238 and the NGB, GB and skGB models were fit on a 2.30GHz Intel Xeon CPU E5-2699 v3.


\section{Results Replication}
A GitHub repository containing the data and code used to produce all results can be found at \url{https://github.com/MikeOMa/MV_Prediction}.

\section{Pre-processing}
\begin{table}[ht]
	\centering
	\begin{tabular}{|c|c|c|}
		\hline
		Name &  unit & Resolution (spatial, temporal)\\
		\hline
		Drifter Speed   & u-v component $cm~s^{-1}$  & irregular, 6 hourly \\
		Wind Speed    & u-v component $m~s^{-1}$ & 0.25 degrees, hourly\\
		Surface wind stress     & u-v component $Pa$ & 0.25 degrees, hourly\\
		Geostrophic Velocity Anomaly     & u-v component $m~s^{-1}$ & 0.25 degrees, daily\\
		Position & location-longitude $degrees$& irregular, 6 hourly\\
		day of year &  $days$& 6 hourly \\
		\hline
	\end{tabular}
	\caption{Summary of data used in the application. Drifter speed is used to define the 2 dimensional response $\mathbf{Y}$. The rest of the variables listed defined the 9 features used for $\mathbf{X}$.}
	\label{tab:Data}
\end{table}

For reproducibility, we now give explicit instructions on how the application dataset used in Section 4 is created.
In its raw form, the drifter dataset is irregularly sampled in time, however the product used here is processed to be supplied on a 6-hourly scale \cite{GDP} and includes location uncertainties. A high uncertainty would be caused by a large gap in the sampling of the raw data or a large satellite positioning error. In our study, we dropped all drifter observations with a positional error higher than $0.5$ in either the longitude or latitudinal coordinates.

The six features we used relate to wind stress, geostrophic velocity, and wind speed, and are all available on longitude-latitude grids, with spatial and temporal resolution specified in Table \ref{tab:Data}; we refer to these as gridded products. Prior to using these products to predict the drifter observations, we spatially interpolated\footnote{The temporal resolutions already match.} the gridded products to the drifter locations. To interpolate the gridded products, an inverse distance weighting interpolation was used. We only used the values at the $n=4$ corners which define the spatial box containing the longitude-latitude location of the drifter location of interest, where the following estimate is used:
\begin{equation}
	\dfrac{\sum_{i=1}^n w_i g_i}{\sum_{i=1}^nw_i},\label{eq:supp_idw}
\end{equation}
where $g_i$ is the gridded value for the $i^{th}$ corner (e.g., a $0.5~m~s^{-1}$ east-west geostrophic velocity at $30^{\circ}$ longitude, $25^\circ$ latitude), $w_i$ is the inverse of the Haversine distance\footnote{The Haversine distance is the greater circle distance between two longitude-latitude pairs.} between the drifter's longitude-latitude and the longitude-latitude location of the gridded value $g_i$.

If two or more of the grid corners which are being interpolated from did not have a value recorded in the gridded product (e.g., if two corners were on a coastline in the case of wind stress and geostrophic velocity), we did not interpolate and treated that point as missing. If only one corner was missing, we simply use $n=3$ in Equation \eqref{eq:supp_idw}.

We low-pass filtered the drifter velocities, wind speed, and wind stress series to remove effects caused by inertial oscillations and tides; this process is similar to previous works \cite{laurindo2017improved}. A critical frequency of 1.5 times the inertial period (inertial period is a function of latitude) was used. Due to there being some missing or previously dropped data due to pre-processing steps, some time series have gaps in time larger than 6 hours. In such cases, we split that time series into individual continuous segments. We applied a fifth-order Butterworth lowpass filter to each continuous segment. If any of these segments were shorter than 18 observations (4.5 days), the whole segment was treated as missing in the dataset. The Butterworth filter was applied in a rolling fashion to account for the changing fashion of the inertial period which defines the critical frequency.

The geostrophic data are available on a daily scale, therefore we decimated all data to daily, only keeping observations at 00:00. No further preprocessing was carried out on the geostrophic velocities (after interpolation) as inertial and tidal motions are not present in the geostrophic velocity product. 

Finally, to remove poorly sampled regions from the dataset, we partitioned the domain into $1^{\circ} \times 1^{\circ}$ longitude-latitude non-overlapping grid boxes. We counted the number of daily observations contained in each box, if this count was lower than 25, then the observations within that box were dropped from the dataset. Functions relating to these steps alongside a link to download the processed data can be found at \url{https://github.com/MikeOMa/MV_Prediction}.

We only considered complete collections for $\mathbf{X}_i,~\mathbf{Y}_i$; if any of the data were recorded as missing in the process explained above, that daily observation was dropped from the dataset. Prior to fitting the models, we scaled the training data $\mathbf{X}_i$ to be in the range $[0,1]$ for all variables, this step was taken to make the neural network fitting more stable.

\section{Multivariate Normal Derivations}
To fit the NGBoost model, we require three elements, the log-likelihood, the derivative of the log-likelihood, and the Fisher information matrix.
We use the parametrization for the multivariate Gaussian given in Section 2.2. 

We shall derive results for the general case where $p$ is the dimension of the data $\mathbf{Y} \in \mathbb{R}^{p}$. We use $Y_i\in \mathbb{R},~i\in \{1,\dots, p\}$ to denote the value of the $i^{th}$ dimension of $\mathbf{Y}$ in this section, a similar notation is used for $\bmu$ and $\mu_i$. The probability density function can be written as:
\begin{equation}
	p(\mathbf{Y} | \boldsymbol{\mu}, \boldsymbol{\Sigma}) = (2\pi)^{-\frac{n}{2}} |\boldsymbol{\Sigma}|^{-{1/2}}\exp\left[-\frac{1}{2} (\mathbf{Y}-\boldsymbol{\mu})\boldsymbol{\Sigma}^{-1}(\mathbf{Y}-\boldsymbol{\mu})\right]. \\
\end{equation}
As mentioned in Section 2, the optimization of $\bmu$ is relatively straightforward as it lives entirely on the real line unconstrained. $\bSigma$ is an $N\times N$ positive definite matrix, which is a difficult constraint. Therefore, optimizing this directly is a challenge. Instead, if we consider the Cholesky decomposition of $\bSigma^{-1} = \mathbf{L}^\top \mathbf{L}$ where $\mathbf{L}$ is an upper triangular matrix, we only require that the diagonal be positive to ensure that $\bSigma$ is positive definite. We choose to model the inverse of $\bSigma$ as in \cite{williams1996using}.

Therefore, we form an unconstrained representation for $\bL$, where we denote $a_{ij}$ as the element in the $i^{th}$ row and $j^{th}$ column as follows:
\begin{equation*}
	a_{ij}=\begin{cases}
		\exp(\nu_{ij}) & \text{If } i=j\\
		\nu_{ij} & \text{If } i<j\\
		0 & \text{Otherwise}
	\end{cases} \text{where } i,j \in \{1,\dots, N\}.
\end{equation*}
Hence, we can fit $\bL$ by doing gradient based optimization on $\{\nu_{ij} \in \mathbb{R} : 1\le i \le j \le n\}$ as all values live on the real line. As an example, for $p=2$ we can fit a multivariate Gaussian using unconstrained optimization through the parameter vector ${\boldsymbol\theta} = \left(\mu_1,~\mu_2, \nu_{11}, \nu_{12}, \nu_{22}\right) \in \mathbb{R}^M$.

As in \cite{williams1996using}, we simplify the parameters to:
\begin{align*}
	z_i &= \mu_i-Y_i  & i \in \{1, \dots, p\}\nonumber\\
	\eta_{i} &= (\mathbf{L}\mathbf{z})_i= \sum_{j=i}^p a_{ij}z_j & i \in \{1, \dots, p\}.
\end{align*}
where we have denoted $\mathbf{z}$ as the column vector of $z_i\in \mathbb R$. Throughout the following derivations we will interchangeably use $\{\bSigma, \bL, a_{ij}, \bmu, \mathbf{z}, z_i, \eta_i\}$ without noting that these parameters have a bijective mapping between them (e.g. $a_{11}=\exp(\nu_{11})$).

Noting that $\log |\boldsymbol{\Sigma}| = \log |\bL^\top \bL|^{-1} = -2\log|\bL| = -2 \log \left(\prod_{i=1}^{p} a_{ii}\right)$,
the negative log-likelihood may be simplified as follows:
\begin{align*}
	-\log p(\mathbf{Y} \given \boldsymbol{\theta} =(\boldsymbol{\mu}, \boldsymbol{\nu})) &= -\dfrac{p}{2}\log(2\pi)+\dfrac{1}{2}\times \log|\boldsymbol{\Sigma}| +  \dfrac{1}{2}(\mathbf{Y}-\boldsymbol{\mu})\boldsymbol{\Sigma}^{-1}(\mathbf{Y}-\boldsymbol{\mu})\\
	&= c-\sum_{i=1}^p \log a_{ii}+\frac{1}{2}\mathbf{z}^\top\bL^\top \bL\mathbf{z}\\
	&= \sum_{i=1}^p \left\{\dfrac{1}{2}\eta_{i}^2-\nu_{ii}\right\} +c,
\end{align*}
where c is a constant independent of $\bmu$ and $\nu_{ij}$. For shorter notation, we will write $l=-\log p(\mathbf{Y} \given \boldsymbol{\theta} =(\boldsymbol{\mu}, \boldsymbol{\nu}))$. 

The first derivatives are stated in \cite{williams1996using} and we also state them here for reference:
\begin{align}
	\dfrac{dl}{d\mu_i} &= \sum_{j=1}^i \eta_j a_{ji} & i\in \{1,\dots, p\}\label{eq:grad_mean} \\
	\dfrac{dl}{d\nu_{ii}} &= \eta_i z_i a_{ii}-1 &i\in \{1,\dots, p\}\label{eq:grad_diag}\\
	\dfrac{dl}{d\nu_{ij}} &= \eta_i z_j &1\le i<j\le p. \label{eq:grad_offdiag}
\end{align}

The final element we need for NGBoost to work is the $M\times M$ Fisher information matrix. We derive the Fisher Information from the following:
\[
\mathcal{I}_{ij} = \Ex \left[\dfrac{d^2l}{d\theta_i d\theta_j} \right], \quad i,~j \in \{1,\dots, M\} .
\]

Note that $\mathcal{I}_{ij}$ is a symmetric matrix, such that $\mathcal{I}_{ij} = \mathcal{I}_{ji}$. For convenience, we denote the entries of the matrix with the subscripts of the parameter symbols, e.g. $\mathcal{I}_{\mu_{i}, \nu_{kq}}$.  We use the letters $i,j,k,q \in \{1, \dots, p\} \subset N$ to index the variables.
The following two expectations are frequently used: $\Ex[z_iz_j]=\boldsymbol{\Sigma}_{ij}$ and $\Ex[z_i]=0$.

The derivatives of $\eta_j$ with respect to the other parameters are repeatedly used, so we give them here:
\begin{align*}
	\dfrac{d\eta_k}{d\mu_i} &= \begin{cases}
		0&  \text{if } k>i\\
		a_{ki}& \text{if } k\le i
	\end{cases} &i,~k\in \{1,\dots, p\}\\
	\dfrac{d\eta_i}{d\nu_{kq}} &= \dfrac{d}{d\nu_{kq}}\sum_{j=i}^n a_{ij}z_j & 1\le i\le p ,\quad 1\le k \le q \le p \\
	&=\begin{cases}
		0 & \text{if } i\ne k\\
		z_q & \text{else if } q > k\\
		a_{ii}z_i & \text{else if } q=k\\
	\end{cases}.
\end{align*}

We start with the Fisher information for $\mu_i, \mu_j$, using the existing derivation of $\frac{dl}{d\mu_k}$ in Equation \eqref{eq:grad_mean}:
\begin{align*}
	\dfrac{d^2l}{d\mu_id\mu_k} &= \dfrac{d}{d\mu_i} \sum_{j=1}^k \eta_ja_{jk} & i,~k\in \{1,\dots, p\}\\
	&= \sum_{j=1}^k a_{ji}a_{jk}\\
	& =\left[\bL^\top \bL\right]_{ik}.
\end{align*}
We therefore have that: 
\begin{equation}
	\mathcal{I}_{\mu_i, \mu_j} = \bSigma_{ij} \quad i,~j\in \{1,\dots, p\}.
	\label{eq:a_mean_mean}
\end{equation}

Next we consider the mean differentiated with respect to $\nu_{kq}$ again starting from Equation \eqref{eq:grad_mean}:
\begin{align*}
	\dfrac{d^2l}{d\mu_id\nu_{kq}} &= \dfrac{d}{d\nu_{kq}} \sum_{j=1}^i \eta_j a_{ji} & i \in \{1,\dots, p\}, ~1\le k\le q\le p \\
	&= \sum_{j=1}^k \left[\eta_j \dfrac{d}{d\nu_{kq}} a_{ji} + a_{ji} \dfrac{d}{d\nu_{kq}} \eta_j\right].\\
\end{align*}
As the expectation of both terms inside the sum is zero for every valid combination of $i,k,q$, we conclude that:
\begin{equation}
	\mathcal{I}_{\mu_{i}, \nu_{kq}} = 0 \quad i\in \{1,\dots, p\} ,~1\le k\le q\le p.
	\label{eq:a_mean_cov}
\end{equation}
Finally, the last elements we require for the Fisher information matrix are the entries for $\nu_{ij}, \nu_{kq}$. 
First, we consider diagonals ($i=j$) w.r.t. all $\nu_{kq}$ with $k\le q$. We start by using Equation \eqref{eq:grad_diag} for $\frac{dl}{d\nu_{ii}}$:
\begin{align*}
	\dfrac{d^2l}{d\nu_{ii} d\nu_{kq}} &= \dfrac{d}{d\nu_{kq}} \eta_i z_ia_{ii} &i\in \{1,\dots, p\} ,~1\le k\le q\le p\\
	&= a_{ii}z_i\dfrac{d}{d\nu_{kq}}\eta_{i}+ \eta_{i}z_i\dfrac{d}{d\nu_{kq}}a_{ii}\\
	&= \begin{cases}
		a_{ii}z_i z_q & \text{if } k=i \text{ and } k<q\\
		a_{ii}^2 z_i z_i& \text{if }i=k=q\\
		0 & \text{if }i \ne k
	\end{cases}
	\\
	&+
	\begin{cases}
		z_i\sum_{j=i}^p a_{ij}z_ja_{ii} & \text{if }i=k=q\\
		0 & \text{Otherwise.}
	\end{cases}
\end{align*}
Taking the expectation we get: 
\begin{equation}
	\mathcal{I}_{\nu_{ii}, \nu_{kq}} = \begin{cases}
		a_{ii}\boldsymbol{\Sigma}_{iq} & \text{if } k=i \text{ and } q>k\\
		a_{ii}^2\boldsymbol{\Sigma}_{ii} + a_{ii}\sum_{j=i}^p a_{ij}\boldsymbol{\Sigma}_{ij} & \text{if }i=k=q\\
		0 & \text{if } i \ne k.
	\end{cases}, \quad i\in \{1,\dots, p\} ,~1\le k\le q\le p 
	\label{eq:a_diag_off_diags}
\end{equation}
Finally, we derive the Fisher information for the off diagonals with respect to the off diagonals ($i<j$ and $k<q$). We start from the expression for $\frac{dl}{v_{ij}}$ in Equation \eqref{eq:grad_offdiag}:
\begin{align*}
	\dfrac{d^2l}{d\nu_{kq} d\nu_{ij}} &= \dfrac{d}{d\nu_{kq}} \eta_i z_j & ~1\le i< j\le p, ~1\le k< q\le p \\
	&= z_j \dfrac{d}{d\nu_{kq}} \eta_i\\
	&= \begin{cases}
		z_jz_q  & \text{if } k=i\\
		0 & \text{if } i\ne k.
	\end{cases}
\end{align*}
Hence, 
\begin{equation}
	\mathcal{I}_{\nu_{ij}, \nu_{kq}} =\begin{cases}
		\Sigma_{jq} & \text{If } k=i \\
		0 & \text{if } k \ne i .
	\end{cases}\quad 1\le i< j\le p, ~1\le k < q\le p.
	\label{eq:a_off_diags}
\end{equation}
Equations \eqref{eq:a_mean_mean}, \eqref{eq:a_mean_cov}, \eqref{eq:a_diag_off_diags}, 
\eqref{eq:a_off_diags} give the full specification of the Fisher Information. 

\subsection{Practical Adjustment}
A small adjustment is made to $\bL$ to improve numerical stability in the implementation. In particular, if the diagonal entries of $\bL$ ($\exp(\nu_{ii})$) get close to zero the resulting matrix $\bSigma^{-1}$ has a determinant extremely close to zero, causing numerical problems to occur when inverting $\bSigma^{-1}$. The fix we propose for this is to add a small perturbation of $10^{-6}$ to the diagonal elements of $\bL$. For most datasets, once the model is fitted, the predicted value for the diagonal elements will be much larger than $10^{-6}$ so this slight adjustment has a negligible effect on the predictions. 

This practical adjustment will become an issue in cases where the determinant of the covariance matrix is very large. As an extreme example, suppose the predicted covariance without the adaptation is $\bSigma = \mathbb{I}_2\times 10^{12}$ where $\mathbb{I}_2$ is the identity matrix. The practical adjustment results in a marginal variance four times larger in this case, which is clearly not negligible. Problems like this can typically be mitigated by scaling the output data $\mathbf{Y}$ (e.g., using a min-max scaling strategy).

\section{Extended Simulation Results}

We now report the metrics introduced in Section 4 for the simulation study in Section 3. These were omitted from the main manuscript due to page length considerations. These are shown in Table \ref{tbl:sim_ext}. We can see that the RMSE metric is very poor for the GB method, which explains the poor NLL and KL divergence metrics for GB. With all methods, NLL and KL divergence show similar patterns, however, if we used the NLL metric it would choose NN for lower values of $N$ over NGB.

We give the results for the unaltered simulation setup of \cite{williams1996using} in Table \ref{tab:williams_sim}. As noted in the main paper, we see that NN does best for $N\in \{500,1000, 3000\}$ and NGB does best for $N\in \{5000,8000, 10000\}$ according to the KL divergence from the truth metric. Moreover, we note GB does not do as poorly as in Table \ref{tbl:sim_ext}. We believe this is because the mean is fit better as can be seen in the RMSE metric. Otherwise, the general patterns seen in both Tables \ref{tbl:sim_ext} and \ref{tab:williams_sim} are very similar.

\begin{table}[ht]
	\caption{Metrics used in Section 4, on the simulation run in Section 3. We also include the KL divergence from Table 1 in the paper for comparison (with one less decimal point shown here to allow the table to fit on the page). The average over the 50 replications is reported. Standard error estimates reported after $\pm$.}
	\label{tbl:sim_ext}
	
	\begin{tabular}{lllllll}
		\toprule
		&       &                   NGB &             Indep NGB &          skGB &             GB &                    NN \\
		Metric & N &                       &                       &               &                &                       \\
		\midrule
		\multirow{6}{*}{KL div} & 500   &  \textbf{0.56 ± 0.02} &           1.63 ± 0.04 &  17.19 ± 0.30 &  126.23 ± 2.58 &           1.28 ± 0.55 \\
		& 1000  &  \textbf{0.26 ± 0.00} &           1.15 ± 0.02 &  17.96 ± 0.27 &  114.11 ± 1.62 &           0.32 ± 0.02 \\
		& 3000  &  \textbf{0.11 ± 0.00} &           0.88 ± 0.01 &  19.61 ± 0.25 &   97.68 ± 1.40 &           0.15 ± 0.00 \\
		& 5000  &  \textbf{0.08 ± 0.01} &           0.88 ± 0.01 &  20.31 ± 0.17 &   90.10 ± 1.29 &           0.13 ± 0.01 \\
		& 8000  &  \textbf{0.05 ± 0.00} &           0.87 ± 0.01 &  20.61 ± 0.17 &   79.01 ± 1.17 &           0.10 ± 0.00 \\
		& 10000 &  \textbf{0.04 ± 0.00} &           0.83 ± 0.01 &  20.55 ± 0.15 &   74.80 ± 1.19 &           0.13 ± 0.00 \\
		\cline{1-7}
		\multirow{6}{*}{NLL} & 500   &           1.35 ± 0.02 &           1.27 ± 0.01 &   2.05 ± 0.01 &    2.49 ± 0.01 &  \textbf{1.05 ± 0.02} \\
		& 1000  &           1.02 ± 0.01 &           1.10 ± 0.01 &   1.92 ± 0.01 &    2.25 ± 0.01 &  \textbf{0.89 ± 0.01} \\
		& 3000  &           0.84 ± 0.01 &           1.02 ± 0.01 &   1.90 ± 0.01 &    2.01 ± 0.01 &  \textbf{0.83 ± 0.01} \\
		& 5000  &  \textbf{0.79 ± 0.01} &           0.98 ± 0.01 &   1.87 ± 0.01 &    1.89 ± 0.01 &           0.81 ± 0.01 \\
		& 8000  &  \textbf{0.78 ± 0.01} &           0.97 ± 0.01 &   1.87 ± 0.01 &    1.80 ± 0.01 &           0.80 ± 0.01 \\
		& 10000 &  \textbf{0.77 ± 0.01} &           0.97 ± 0.01 &   1.88 ± 0.01 &    1.75 ± 0.01 &           0.82 ± 0.01 \\
		\cline{1-7}
		\multirow{6}{*}{RMSE} & 500   &           0.66 ± 0.00 &           0.65 ± 0.00 &   0.65 ± 0.00 &    1.91 ± 0.01 &  \textbf{0.65 ± 0.00} \\
		& 1000  &           0.63 ± 0.00 &           0.63 ± 0.00 &   0.63 ± 0.00 &    1.85 ± 0.01 &  \textbf{0.62 ± 0.00} \\
		& 3000  &           0.63 ± 0.00 &           0.62 ± 0.00 &   0.63 ± 0.00 &    1.76 ± 0.01 &  \textbf{0.62 ± 0.00} \\
		& 5000  &           0.62 ± 0.00 &           0.62 ± 0.00 &   0.62 ± 0.00 &    1.70 ± 0.01 &  \textbf{0.62 ± 0.00} \\
		& 8000  &           0.62 ± 0.00 &  \textbf{0.62 ± 0.00} &   0.62 ± 0.00 &    1.65 ± 0.01 &           0.62 ± 0.00 \\
		& 10000 &           0.62 ± 0.00 &  \textbf{0.62 ± 0.00} &   0.62 ± 0.00 &    1.64 ± 0.01 &           0.62 ± 0.00 \\
		\cline{1-7}
		\multirow{6}{*}{90\% PR area} & 500   &           1.90 ± 0.02 &           2.59 ± 0.03 &   4.27 ± 0.05 &    8.57 ± 0.10 &           3.02 ± 0.07 \\
		& 1000  &           1.97 ± 0.01 &           2.52 ± 0.02 &   4.54 ± 0.04 &    7.74 ± 0.07 &           2.64 ± 0.02 \\
		& 3000  &           2.22 ± 0.01 &           2.60 ± 0.01 &   4.96 ± 0.03 &    7.21 ± 0.06 &           2.58 ± 0.02 \\
		& 5000  &           2.30 ± 0.01 &           2.65 ± 0.01 &   5.09 ± 0.02 &    6.89 ± 0.05 &           2.61 ± 0.03 \\
		& 8000  &           2.36 ± 0.01 &           2.69 ± 0.01 &   5.14 ± 0.02 &    6.49 ± 0.05 &           2.65 ± 0.03 \\
		& 10000 &           2.38 ± 0.01 &           2.69 ± 0.01 &   5.19 ± 0.02 &    6.37 ± 0.05 &           2.63 ± 0.03 \\
		\cline{1-7}
		\multirow{6}{*}{90\% PR cov} & 500   &           0.76 ± 0.00 &           0.83 ± 0.00 &   0.81 ± 0.00 &    0.84 ± 0.00 &           0.88 ± 0.00 \\
		& 1000  &           0.80 ± 0.00 &           0.85 ± 0.00 &   0.83 ± 0.00 &    0.86 ± 0.00 &           0.89 ± 0.00 \\
		& 3000  &           0.85 ± 0.00 &           0.86 ± 0.00 &   0.85 ± 0.00 &    0.88 ± 0.00 &           0.89 ± 0.00 \\
		& 5000  &           0.87 ± 0.00 &           0.87 ± 0.00 &   0.86 ± 0.00 &    0.89 ± 0.00 &           0.90 ± 0.00 \\
		& 8000  &           0.87 ± 0.00 &           0.88 ± 0.00 &   0.87 ± 0.00 &    0.90 ± 0.00 &           0.90 ± 0.00 \\
		& 10000 &           0.88 ± 0.00 &           0.88 ± 0.00 &   0.86 ± 0.00 &    0.90 ± 0.00 &           0.90 ± 0.00 \\
		\bottomrule
	\end{tabular}
\end{table}

\begin{table}[ht]
	\centering
	\caption{Same as Table \ref{tbl:sim_ext} except we use the same simulation as \cite{williams1996using}, i.e. without adding the $+x$ and $-x^2$ terms of equation (3) in the main paper. Note that the difference between NGB and GB or NN is not as large as in the original table. All values rounded to two decimal places.}
	\label{tab:williams_sim}
	\begin{tabular}{lllllll}
		\toprule
		&       &                   NGB &    Indep NGB &          skGB &                    GB &                    NN \\
		Metric & N &                       &              &               &                       &                       \\
		\midrule
		\multirow{6}{*}{KL div} & 500   &           0.96 ± 0.03 &  2.27 ± 0.06 &  19.84 ± 0.39 &           2.16 ± 0.06 &  \textbf{0.40 ± 0.04} \\
		& 1000  &           0.46 ± 0.01 &  1.44 ± 0.03 &  20.08 ± 0.29 &           1.28 ± 0.03 &  \textbf{0.26 ± 0.04} \\
		& 3000  &           0.18 ± 0.01 &  1.06 ± 0.02 &  20.43 ± 0.19 &           0.75 ± 0.01 &  \textbf{0.13 ± 0.02} \\
		& 5000  &  \textbf{0.11 ± 0.00} &  0.96 ± 0.02 &  20.65 ± 0.20 &           0.65 ± 0.01 &           0.13 ± 0.02 \\
		& 8000  &  \textbf{0.07 ± 0.00} &  0.91 ± 0.01 &  20.75 ± 0.17 &           0.55 ± 0.01 &           0.11 ± 0.02 \\
		& 10000 &  \textbf{0.06 ± 0.00} &  0.92 ± 0.02 &  21.10 ± 0.18 &           0.54 ± 0.01 &           0.13 ± 0.02 \\
		\cline{1-7}
		\multirow{6}{*}{NLL} & 500   &           1.24 ± 0.01 &  1.25 ± 0.01 &   1.96 ± 0.01 &           1.27 ± 0.01 &  \textbf{0.93 ± 0.02} \\
		& 1000  &           1.03 ± 0.01 &  1.12 ± 0.01 &   1.91 ± 0.01 &           1.12 ± 0.01 &  \textbf{0.86 ± 0.01} \\
		& 3000  &           0.86 ± 0.01 &  1.02 ± 0.01 &   1.88 ± 0.01 &           0.98 ± 0.01 &  \textbf{0.82 ± 0.01} \\
		& 5000  &           0.82 ± 0.01 &  1.00 ± 0.01 &   1.86 ± 0.01 &           0.94 ± 0.01 &  \textbf{0.81 ± 0.01} \\
		& 8000  &  \textbf{0.78 ± 0.01} &  0.97 ± 0.01 &   1.86 ± 0.01 &           0.90 ± 0.01 &           0.80 ± 0.01 \\
		& 10000 &  \textbf{0.78 ± 0.01} &  0.97 ± 0.01 &   1.86 ± 0.01 &           0.90 ± 0.01 &           0.81 ± 0.01 \\
		\cline{1-7}
		\multirow{6}{*}{RMSE} & 500   &           0.64 ± 0.00 &  0.64 ± 0.00 &   0.63 ± 0.00 &  \textbf{0.63 ± 0.00} &           0.64 ± 0.00 \\
		& 1000  &           0.63 ± 0.00 &  0.63 ± 0.00 &   0.63 ± 0.00 &  \textbf{0.62 ± 0.00} &           0.63 ± 0.00 \\
		& 3000  &           0.62 ± 0.00 &  0.62 ± 0.00 &   0.62 ± 0.00 &  \textbf{0.62 ± 0.00} &           0.62 ± 0.00 \\
		& 5000  &           0.61 ± 0.00 &  0.61 ± 0.00 &   0.61 ± 0.00 &  \textbf{0.61 ± 0.00} &           0.62 ± 0.00 \\
		& 8000  &           0.62 ± 0.00 &  0.62 ± 0.00 &   0.62 ± 0.00 &  \textbf{0.61 ± 0.00} &           0.62 ± 0.00 \\
		& 10000 &           0.62 ± 0.00 &  0.62 ± 0.00 &   0.62 ± 0.00 &  \textbf{0.62 ± 0.00} &           0.62 ± 0.00 \\
		\cline{1-7}
		\multirow{6}{*}{90\% PR area} & 500   &           2.03 ± 0.02 &  2.61 ± 0.03 &   4.68 ± 0.06 &           2.72 ± 0.03 &           2.79 ± 0.05 \\
		& 1000  &           2.11 ± 0.02 &  2.53 ± 0.02 &   4.86 ± 0.05 &           2.65 ± 0.02 &           2.79 ± 0.04 \\
		& 3000  &           2.31 ± 0.01 &  2.65 ± 0.01 &   5.10 ± 0.03 &           2.69 ± 0.01 &           2.71 ± 0.03 \\
		& 5000  &           2.35 ± 0.01 &  2.67 ± 0.01 &   5.14 ± 0.02 &           2.70 ± 0.01 &           2.69 ± 0.02 \\
		& 8000  &           2.41 ± 0.01 &  2.70 ± 0.01 &   5.20 ± 0.02 &           2.71 ± 0.01 &           2.73 ± 0.03 \\
		& 10000 &           2.42 ± 0.01 &  2.72 ± 0.01 &   5.24 ± 0.02 &           2.72 ± 0.01 &           2.70 ± 0.03 \\
		\cline{1-7}
		\multirow{6}{*}{90\% PR cov} & 500   &           0.78 ± 0.00 &  0.83 ± 0.00 &   0.84 ± 0.00 &           0.84 ± 0.00 &           0.88 ± 0.00 \\
		& 1000  &           0.81 ± 0.00 &  0.84 ± 0.00 &   0.85 ± 0.00 &           0.85 ± 0.00 &           0.89 ± 0.00 \\
		& 3000  &           0.86 ± 0.00 &  0.87 ± 0.00 &   0.86 ± 0.00 &           0.88 ± 0.00 &           0.89 ± 0.00 \\
		& 5000  &           0.86 ± 0.00 &  0.87 ± 0.00 &   0.87 ± 0.00 &           0.88 ± 0.00 &           0.90 ± 0.00 \\
		& 8000  &           0.88 ± 0.00 &  0.88 ± 0.00 &   0.87 ± 0.00 &           0.89 ± 0.00 &           0.90 ± 0.00 \\
		& 10000 &           0.87 ± 0.00 &  0.88 ± 0.00 &   0.87 ± 0.00 &           0.89 ± 0.00 &           0.90 ± 0.00 \\
		\bottomrule
	\end{tabular}
\end{table}

\end{document}